# Alife Model of Evolutionary Emergence of Purposeful Adaptive Behavior[1]

Mikhail S. Burtsev[1], Vladimir G. Red'ko[2], Roman V. Gusarev[3]

Keldysh Institute of Applied Mathematics, 4 Miusskaya sq., Moscow RU-125047, Russia
[1]`mr.bur@beep.ru`, [2]`redko@keldysh.ru`, [3]`gusarer@mail.ru`

**Abstract.** The process of evolutionary emergence of purposeful adaptive behavior is investigated by means of computer simulations. The model proposed implies that there is an evolving population of simple agents, which have two natural needs: energy and reproduction. Any need is characterized quantitatively by a corresponding motivation. Motivations determine goal-directed behavior of agents. The model demonstrates that purposeful behavior does emerge in the simulated evolutionary processes. Emergence of purposefulness is accompanied by origin of a simple hierarchy in the control system of agents.

## 1 Introduction

The purposefulness is very non-trivial feature of intelligent animal behavior. What are the particularities of purposeful adaptive behavior? How could goal-directed behavior emerge? We try to investigate these questions by means of computer simulations.

The main notion that we use to characterize purposefulness is that of *motivation*. Motivation is quantitative characteristics describing the intention of an agent to reach a goal.

The key assumptions of our model are as follows:

1. There are agents, which have two natural needs (the need of energy and the need of reproduction).
2. The population of agents evolves in the simple environment, where patches of grass (agent's food) grow. The agents receive some information from their environment and perform some actions. Agents can move, eat grass, rest and mate with each other. Mating results in birth of new agents. An agent has an internal energy resource; the resource is increased during eating. Performing an action, the agent spends its resource. When the resource of the agent goes to zero, the agent dies.
3. Any need of an agent is characterized by a quantitative parameter (motivation parameter) that determines the motivation to reach a corresponding purpose. E.g., if the energy resource of an agent is small, there is the motivation to find food and to replenish the energy resource by eating.

---

[1] URL: http://www.keldysh.ru/mrbur-web/publ/ecal01full.pdf

4. The agent behavior is controlled by a neural network, which has special inputs from motivations. If there is a certain motivation, the agent can search for a solution to satisfy the need according to the motivation. This type of behavior can be considered as purposeful (there is the purpose to satisfy the need).
5. The population of agents evolves. The main mechanism of the evolution is the formation of genomes of new agents with the aid of formal genetic operators. A genome codes the synaptic weights of the agent's neural network.

The following works form the background of our research:

- the model of PolyWord by L. Yaeger [1],
- the analysis of interactions between learning and evolution by D. Ackley and M. Littman [2],
- the models of evolving neural networks by S. Nolfi and D. Parisi [3],
- the model of motivation by L.E. Tsitolovsky [4],
- the investigations of motivationally autonomous animat "MonaLysa" by J.Y. Donnart and J.-A. Meyer [5],
- the analysis of the roots of motivation by C. Balkenius [6].

The agents in our model are similar (but simpler) to "organisms" in PolyWord [1]. The idea of special evolving neural network, that is intended to give rise to agent's goal [2], is to some extent analogous to special neural network inputs, controlled by motivations, in our model. The importance of the notion of motivation, underlined by the authors of the works [4-6], stimulates the introduction of this concept into our model.

## 2 The Model

### 2.1 Overview of Environment and Agents

The environment is a linear one-dimensional set of cells (Fig. 1). We assume that only a single agent can occupy any cell.

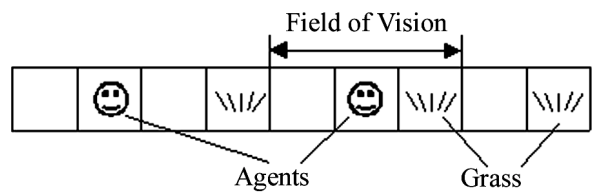

**Fig. 1.** Agents in the one-dimensional cellular environment.

The time is discrete. At any iteration, each agent executes exactly one action. The set of possible actions of agents is the following: 1) resting; 2) moving to a neighboring cell (to the left or to the right); 3) jumping (over several cells into random direction); 4) eating; 5) mating.

The grass patches appear randomly and grow certain time at some cells of the environment.

The agents are "short-sighted". This means that any agent views the situation only in three cells: in its own cell and in two neighboring cells. We designate these 3 cells as "field of vision" of an agent (Fig. 1).

We introduce two quantitative parameters, corresponding to the agents' needs:
1. Motivation to search the food $M_E$, which corresponds to the need of energy;
2. Motivation to mating $M_R$, which corresponds to the need of reproduction.

Motivations are defined as follows (Fig. 2):

$$M_R = \min\left\{\frac{R}{R_1},\ 1\right\},\ M_E = \max\left\{\frac{R_0 - R}{R_0},\ 0\right\},$$

where $R_0$ is some "optimal" value of energy resource, $R_1$ is the value of energy resource, which is the most appropriate for reproduction.

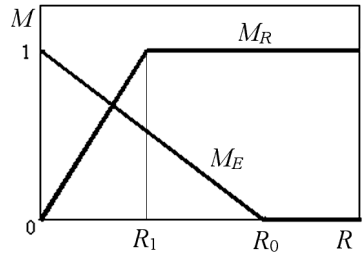

**Fig. 2.** Dependence of motivations on energy resource of an agent.

### 2.2 Interactions between Agents and Environment

**Eating.** An agent is able to determine the presence of food in its "field of vision". An agent can eat the food in its own cell. We assume that if an agent "decides" to eat, it eats all amount of food in the cell during one moment of time. While eating, the agent replenishes its energy resource.

**Mating.** Executing action "mating", an agent loses some energy in any case. If mating actions of two agents in the neighboring cells have been coordinated, the child is born and both parents transfer some amount of energy resource to the offspring. Initial energy of the child equals to the energy obtained from parents. The genome of the offspring is a recombination of parent's genomes (slightly modified by mutations). The new agent is placed into one of two possible cells, neighboring to the pair of parent's cells. If there is no free cell in the neighborhood around the parents, then the offspring dies off. The latter implies that the world is too densely populated, so there is insufficient "living space" for the child.

### 2.3 Neural Network of an Agent

The neural network of an agent controls its behavior. We suppose that the neural network includes one layer of neurons. The neurons receive signals from external and internal environment via sensory inputs. There are full interconnections between sensory inputs and neurons: each neuron is connected to all inputs. The outputs of neurons determine agent's actions.

We assume that an agent receives from external environment the following information:
1. Food existence in every cell of field of vision.
2. Other agent presence in the left cell.
3. Other agent presence in the right cell.
4. Motivation to mate of agents in the left and in the right cells.

(Here we imply that the motivation to mate determines the color of an agent, and its neighbors register this color. This property is similar to that of blue color of agents in the PolyWord [1].)

So, we have 3+1+1+2=7 parameters, characterizing the external environment.

Additionally, we assume that an agent receives two motivation parameters $M_E$, $M_R$ from the internal environment.

Thus, the neural network of an agent has 9 sensory inputs.

The neurons determine the agent's actions. These actions are as follows:
1. To be at rest.
2. Moving (to the left or to the right) to a neighboring cell.
3. Jumping randomly over several cells.
4. Eating.
5. Mating with one of two possible neighbors.

Taking into account that actions 2 and 5 have two alternatives, we see that there are 1+2+1+1+2= 7 different actions, the agent can perform. Correspondingly, there are 7 neurons.

The neurons are assumed to have typical logistic activation function.

We assume that at the given moment of time the agent accomplish only one action, namely, the action, corresponding to that neuron, which has maximal output signal.

Since inputs and neurons have all possible synaptic interconnections, there are 9x7 = 63 synaptic weights in the neural network.

### 2.4 Scheme of the Evolution

The scheme of the evolution is implemented in the following way. We assume that a genome of an agent *S* codes synaptic weights of the agent's neural network:

$$S = (S_1, S_2, \ldots, S_N), N = 63.$$

Each synaptic weight $S_i$ is represented by real number and considered as a gene of the genome. When a new agent is being born, its genome is created in the following manner: 1) a uniform recombination of parent's genomes is formed, 2) this recombined genome is subjected to small mutations.

During the uniform recombination, the genes $S_i$ of the offspring genome are taken from any parent randomly for each gene. During mutations, random values $z$, uniformly distributed in the interval [$-p_m$, $p_m$], are added to the value of every gene:

$$S_i \rightarrow S_i + z_i, i = 1,2, \ldots, N,$$

where $p_m$ is the mutation intensity.

## 3. Computer Simulations

### 3.1 Simulations Parameters

To analyze the influence of motivations on behavior of agents, we performed two series of simulations. In the first series, the agents had motivations (the motivations were introduced as described above); and in the second series, the agents had no motivations (the inputs from motivations were artificially suppressed by means of special choice of parameters $M_E$, $M_R$). In order to analyze the influence of food amount in the external environment on population behavior, the simulations in both series were performed for the several probabilities of grass appearance in cells.

The size of artificial world in all simulations was equal to 900 cells.

We defined some reasonable agent's physiology by choosing parameters, which determine its energy consumption. These parameters were defined in the following way:
1. when an agent accomplishes the action "resting", it losses some minimal amount of energy: $\Delta R_r = - \Delta R_{min} = -r$;
2. the action "eating" takes two times greater energy amount: $\Delta R_e = - 2r$;
3. for the action "moving", the energy consumption is two times more greater: $\Delta R_{mv} = - 4r$;
4. the actions "jumping" and "mating" take five times more energy as compared with "moving"(we set that the jumping agent moves over five cell to the left or to the right): $\Delta R_j = - 20r$; $\Delta R_{mt} = - 20r$;
5. the amount of energy transferred from parents to a child was set to be 1000 times greater than for "resting": $\Delta R_{ch} = 1000r$.

The parameters characterizing motivations were determined in the following manner: $R_0 = 10^4 r$, and $R_1 = 0.5 R_0$.

The amount of energy that an agent obtains during eating one portion of grass was set to be equal to $0.02 R_0$. Lifetime of a grass patch was equal to 20 iterations of time. The agents in an initial population had energy resource of the order of $R_0$.

These parameters ensured that the lifetime of agents was typically rather long (100-10000 iterations), thus simulated evolution was rather gradual.

The mutation intensity $p_m$ was equal to 0.05. The number of agents in an initial population was equal to 200.

Food amount in the environment was characterized by a parameter $P_g$. At any iteration, a patch of grass was appeared in any cell with probability $P_g$.

All agents of the initial population had the same synaptic weights of neural networks. These weights determined some reasonable initial instincts of agents.

The first instinct was the instinct of food replenishment. This instinct ensured two types of actions: 1) if an agent sees a grass in its own cell, it eats this grass, 2) if an agent sees a grass in a neighboring cell, it moves into this cell.

The second instinct was the instinct of reproduction. This instinct implies that if an agent sees another agent in one of the neighboring cells, it tries to mate with this neighbor.

In addition to these main instincts, the agents were provided with the instinct of "fear of tightness": if an agent sees two agents in the both neighboring cells, it jumps.

The synaptic weights from motivational inputs in neural network were equal to zero for all of agents in initial population.

### 3.2 Results

The main quantitative characteristic that we used in order to describe the quality of an evolutionary process was the total number of agents in population $N$. We obtained the dependencies $N(t)$ on time $t$ for both series of experiments: for population of agents with motivations and for population of agents without motivations. We also analyzed evolutionary dynamics of agents actions and registered a statistics of the synaptic weights during a process of evolution.

Examples of the dependencies $N(t)$ are shown in Fig. 3. At small amount of food (Fig. 3a), both populations of agents (with and without motivations) die out – the amount of food is not enough to support energy consumption needed for agent actions. At an intermediate amount of food (Fig. 3b), the population of agents without motivations dies out, whereas the population of agents with motivations is able to find a "good" living strategy and survives. At large amount of food (Fig. 3c) both populations survive, however, the population with motivations finds better neural network control system for its agents, which assure the larger final population.

Thus, neural network inputs from internal motivations provide an opportunity for the population to find better control system for agents in the course of evolutionary search.

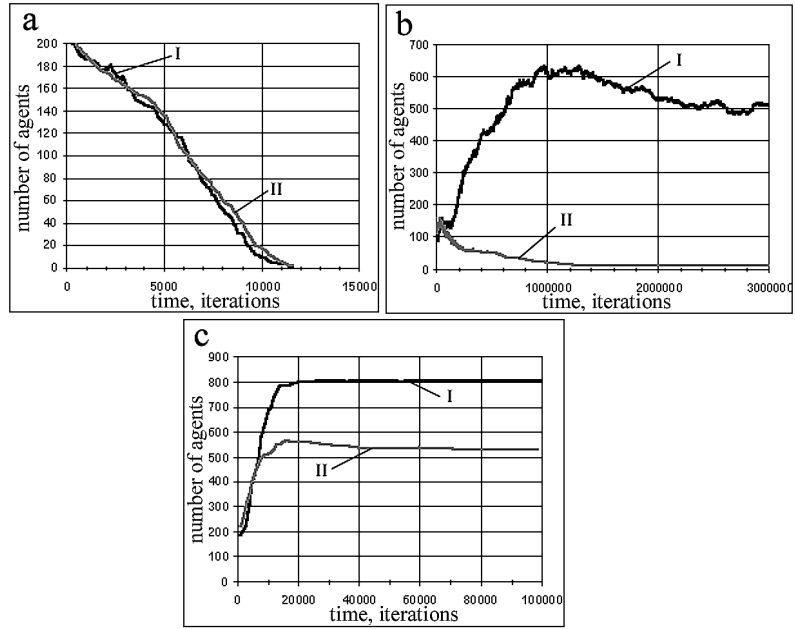

**Fig. 3.** Dependencies of number of agents in population with motivations (I) and without motivations (II) on time for different probabilities of grass appearance, $P_g$: a) $P_g = 1/2000$, b) $P_g = 1/200$, c) $P_g = 1/20$.

### 3.3 Analysis

We performed detailed analysis of agents' actions evolution for population with and without motivations. Based on this analysis, we interpreted behavioral control of agents.

The scheme of behavioral control of an agent *without motivations* that was discovered by evolution is shown at Fig. 4. This scheme consists of three rules, which are used by the agent during its life.

The first rule says that if an agent sees a grass patch, it seeks to eat this food. Namely, it eats food, if the food is in its own cell, or goes to grassy neighboring cell and eats food at the next moment of time.

The second rule says that if the agent sees a neighbor, it makes mating action, trying to give birth to an offspring.

These two rules are just instincts, which we forced upon the agents of an initial population. The evolution confirmed that they are useful and adaptive.

The third rule says that if the agent doesn't see anything in its field of vision, it decides to rest. This rule was discovered by evolution, and, of course, the rule has a certain adaptive value.

It is obvious that such agent behavior is determined by current state of external environment only. These three rules can be considered as simple reflexes.

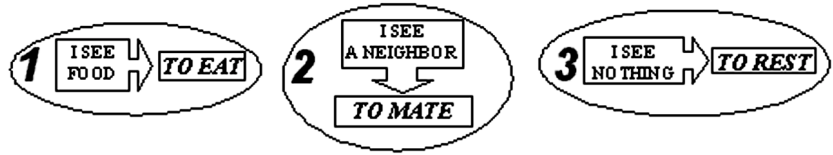

**Fig. 4.** Scheme of behavioral control of agents without motivations.

Let's consider the control system of the agent *with motivations*. The analysis of simulations demonstrates that the control scheme of the agent with motivations can be represented as a hierarchical system. Three rules described above constitute the lower level of the control system. The second level is due to motivations. This hierarchical control system works in the following manner (Fig. 5).

If the energy resource of the agent is low, the motivation to search food is large, and the motivation to mating is small, so the agent uses only two of mentioned rules, the first and the third – the mating is suppressed. If the energy resource of the agent is high, the motivation to mating is turned on, and so the agent seeks to mate – the second and the third rules govern mainly the agent behavior, however, sometimes the first rule works too.

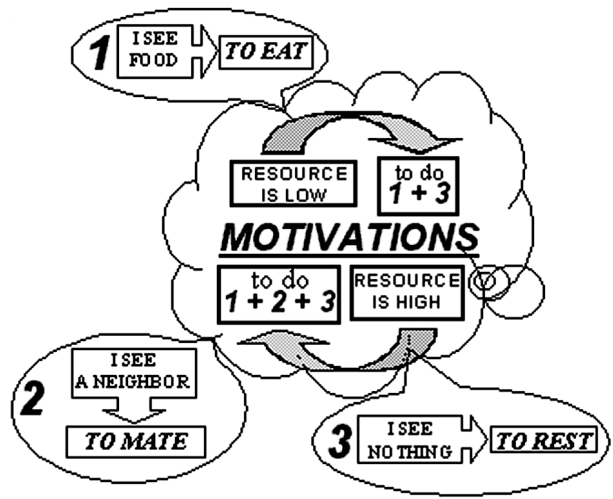

**Fig. 5.** Scheme of behavioral control of agents with motivations.

So, the transition from the scheme of control without motivations (Fig. 4) to the scheme with motivations (Fig. 5) can be considered as the emergence of a new level of hierarchy in the control system of the agents. This transition is analogous to the metasystem transition from simple reflexes to complex reflex in the Metasystem Transition Theory by V. Turchin [7].

## 4   Conclusion

Thus, the model of evolutionary emergence of purposeful adaptive behavior has been developed. The model demonstrates that simple hierarchical control system, where simple reflexes are controlled by motivations, can emerge in evolutionary processes, and this hierarchical system is more effective as compared to behavioral control governed by means of simple reflexes only.